\documentclass[journal]{IEEEtran}

\usepackage{cite}

\ifCLASSINFOpdf
\else
\fi

\usepackage[cmex10]{amsmath}
\usepackage{color}
\usepackage{amsmath}
\usepackage[amsmath,thmmarks]{ntheorem}
\usepackage{amsfonts}
\usepackage{amssymb}

\usepackage{color,soul}

\usepackage{graphicx}
\usepackage[tight,footnotesize]{subfigure}
\usepackage{setspace}
\usepackage{balance}
 \usepackage{mathrsfs}
 \usepackage{amsmath}

 \usepackage{amssymb}
\usepackage{stfloats}
\usepackage{verbatim}
\usepackage[ruled]{algorithm2e}
\usepackage[amsmath,thmmarks]{ntheorem}
\usepackage{bm}
\usepackage[justification=centering]{caption}

\begin{document}

\title{Adaptive Federated Learning and Digital Twin for Industrial Internet of Things  \\
{\footnotesize \textsuperscript{} }
\thanks{}
}

\author{Wen~Sun,~\IEEEmembership{~Member,~IEEE,}
	~Shiyu~Lei,~\IEEEmembership{~Student Member,~IEEE,}
	~Lu~Wang,~\IEEEmembership{~Student Member,~IEEE,}
	~Zhiqiang~Liu,~\IEEEmembership{~Member,~IEEE,}
	and~Yan~Zhang,~\IEEEmembership{~Fellow,~IEEE}
	
	\IEEEcompsocitemizethanks{W. Sun and Z. Liu are with the School of Cybersecurity, Northwestern Polytechnical University, 127 West Youyi Road, Xi'an Shaanxi, 710072, China (e-mail: sunwen@nwpu.edu.cn, zqliu@nwpu.edu.cn). S. Lei and L. Wang are with the School of Cyber Engineering, Xidian University, No. 2 South Taibai Road, Xi'an Shaanxi, 710071, China (e-mail: leishiyu\_xd@163.com, wangl\_xd@163.com).  Y. Zhang is with University of Oslo, Norway; and also with Simula  Metropolitan Center for Digital Engineering, Norway (e-mail: yanzhang@ieee.org).
	
	}
}

\maketitle

\begin{abstract}
\textbf{
Industrial Internet of Things (IoT) enables distributed intelligent services varying with the dynamic and realtime industrial environment to achieve Industry 4.0 benefits. In this paper, we consider a new architecture of digital twin empowered Industrial IoT where digital twins capture the characteristics of industrial devices to assist federated learning. Noticing that digital twins may bring estimation deviations from the actual value of device state, a trusted based aggregation is proposed in federated learning to alleviate the effects of such deviation. We adaptively adjust the aggregation frequency of federated learning based on Lyapunov dynamic deficit queue and deep reinforcement learning, to improve the learning performance under the resource constraints. To further adapt to the heterogeneity of Industrial IoT, a clustering-based asynchronous federated learning framework is proposed. Numerical results show that the proposed framework is superior to the benchmark in terms of learning accuracy, convergence, and energy saving.}
\end{abstract}

\begin{IEEEkeywords}
\textbf{digital twin, federated learning, asynchronous, learning efficiency, communication efficiency.}
\end{IEEEkeywords}

\section{Introduction}
Industrial Internet of Things (IoT), as an extension use of IoT, interconnects numerous industrial devices, analytical sections, and people at work. Through exquisite cooperation of machinery devices, Industrial IoT enables automatic manufacturing applications ranging from automotive to agriculture, and finally realizing Industry 4.0 [1]. The success of Industrial IoT hinges on dynamic perception and intelligent decision, which is difficult to capture due to the heterogeneous Industrial IoT devices and the complex industrial environment [2]. 

Digital twin (DT), as an emerging digitalization technology, offers a feasible solution to capture the dynamic and complex industrial environment [3]. It creates virtual objects in the digital space through software definition, accurately mapping entities in the physical space in terms of status, features and evolution. Its excellent state awareness and real-time analysis greatly assist decision-making and execution. However, DTs are data-driven and a decision making in Industrial IoT usually needs to be supported by a large amount of data distributed across a variety of training devices, while “data islands” exist in most data sources due to competition, privacy and security issues [4]. In reality, it is almost impossible to integrate data scattered in various devices. The dual challenges of privacy protection and big data present an opportunity to develop new technologies that aggregate and model the data under strict prerequisites.

Federated learning turns out to be a powerful weapon to enable distributed advanced analysis in Industrial IoT [5]. Federated learning is a model training technology that data owners can perform model training locally without sharing their data, which helps ensure privacy and reduce communication costs. There have been works on designing advanced federated learning algorithms to achieve better learning performance including privacy preservation and learning efficiency [6] [7]. To address the limited communication bandwidth, Yang \emph{et al.} [7] proposed a fast global aggregation scheme by device selection and beamforming design. The problem is solved using an optimization problem with global convergence guarantees. In order to improve the learning effect, Lu \emph{et al.} [8] introduce a learning framework of asynchronous mode, which accelerated the convergence speed of learning, but the point-to-point communication mode caused a great communication burden.

Existing efforts [6-8] do not take into account the dynamic Industrial environment and the adaptive federated learning architecture accordingly. In fact, the frequency and timing of aggregation should be carefully designed in federated learning, as the gain of global aggregation is non-linear and the network environment, e.g. the channel state, is time-varying during the federated learning process. It is also noted that for heterogeneous Industrial IoT scenario, the straggler effect makes the synchronized federated learning suffer an unbearable learning delay. To address these issues, we study adaptive calibration of the global aggregation frequency based on knowledge from DTs to improve training efficiency under resource constraints.  Furthermore, we propose an  asynchronous federated learning framework based on node clustering. The contributions of this paper can be summarized as follows.

\begin{itemize}
\item We introduce DTs for Industrial IoT, which  map
the operating state and behavior of devices to a digital world in real time.  By considering the deviation of the DT from the true value in the trust-weighted aggregation strategy,  the contribution of devices to the global aggregation of federated learning is quantified,  which enhances the reliability and accuracy of learned models.
  \item Based on  \emph{Deep  Q Network} (\emph{DQN}), we develop an adaptive calibration of global aggregation frequency  to minimize the loss function of federated learning under a given resource budget,  enabling the dynamic tradeoff between computing energy and communication energy in  time-varying communication environments.
  \item To  further adapt to heterogeneous Industrial IoT, we propose an asynchronous federated learning framework to eliminate the straggler effect by clustering nodes, and improve the learning efficiency through appropriate  time-weighted inter-cluster aggregation strategy. The aggregation frequencies of different clusters are determined by the adaptive frequency calibration based on DQN. Numerical results show that the proposed scheme is superior to the benchmark scheme in terms of learning accuracy, convergence rate and energy saving.



\end{itemize}

The organization of the remaining paper is as follows. In
Section II, we overview the related works. The DT-based system model is introduced in Section III. The frequency of global aggregation and an asynchronous federated learning framework are carefully designed based on \emph{DQN} in Section IV.  The simulation results of  the proposed scheme performance are provided in  Section V. Finally,
Section VI concludes the paper.

\section{Related Work}

Existing works of federated learning mainly focuse on  update architecture, aggregation strategy and frequency aggregation.

\textbf{Update architecture.} 
Most of the current algorithms use synchronous architecture, such as FedAvg [5].  However, the synchronous architecture makes the duration of training limited to slower nodes and not suitable for scenarios where the node resources are heterogeneous, i.e., the straggler effect.
A few studies have also considered asynchronous learning, e.g.,  \emph{Lu et al.} \cite{o2} allow asynchronous updates through a random distributed update scheme, but it does cause out-of-order communication between nodes, which is indeed a huge communication burden. In addition, \emph{Fadlullah  et al.} \cite{b2} studied the asynchronous update of weights, which supports shallow parameters to be updated more frequently in an asynchronous manner but does not eliminate the straggler effect.

\textbf{Aggregation strategy.}  
The current research has explored the influence of factors such as data size, computing power, and reputation value on aggregation [10-12],  which are closely related to application scenarios. To capture the relationship between non-IID and unbalanced data, FedAvg  weighted the data amount of the training nodes. In a resource-constrained scenario, \emph{Nishio et al.} \cite{b6} comprehensively consider the data resources, computing power, and wireless channel conditions of heterogeneous mobile devices to accelerate convergence. \emph{Pandey et al.} \cite{b7} introduce communication efficiency to measure the reliability of equipment, and establish a probability model to calculate the corresponding aggregation weight.
Obviously, finding out the important factors that affect the training effect under the specific application scenario is the top priority in solving the aggregation challenge of federated learning.

\textbf{Aggregation frequency.} 
The adaptive calibration of the global aggregation frequency is  beneficial to improve the scalability and efficiency of federated learning \cite{20}. To better adapt to dynamic resource constraints, \emph{Wang et al.} \cite{21} proposed an adaptive global aggregation scheme to change the global aggregation frequency to improve training efficiency under resource constraints. Similarly, \emph{Tran et al.} [17] captured the trade-off between computation and communication delay determined by learning accuracy, 
and defined  an optimization problem about the global optimal learning time which can be decomposed into three convex subproblems to solve.

In general, there are still some issues to be solved in federated learning, i.e., the  straggler  effect is not completely eliminated, the adaptive aggregation frequency in Industrial IoT. In this paper, we combine DTs and \emph{DQN} to adaptively reduce energy consumption and design an asynchronous federated learning framework to eliminate the straggler effect.

\section{System Model}

\begin{figure*}[t]
\centering
\includegraphics[scale=0.5]{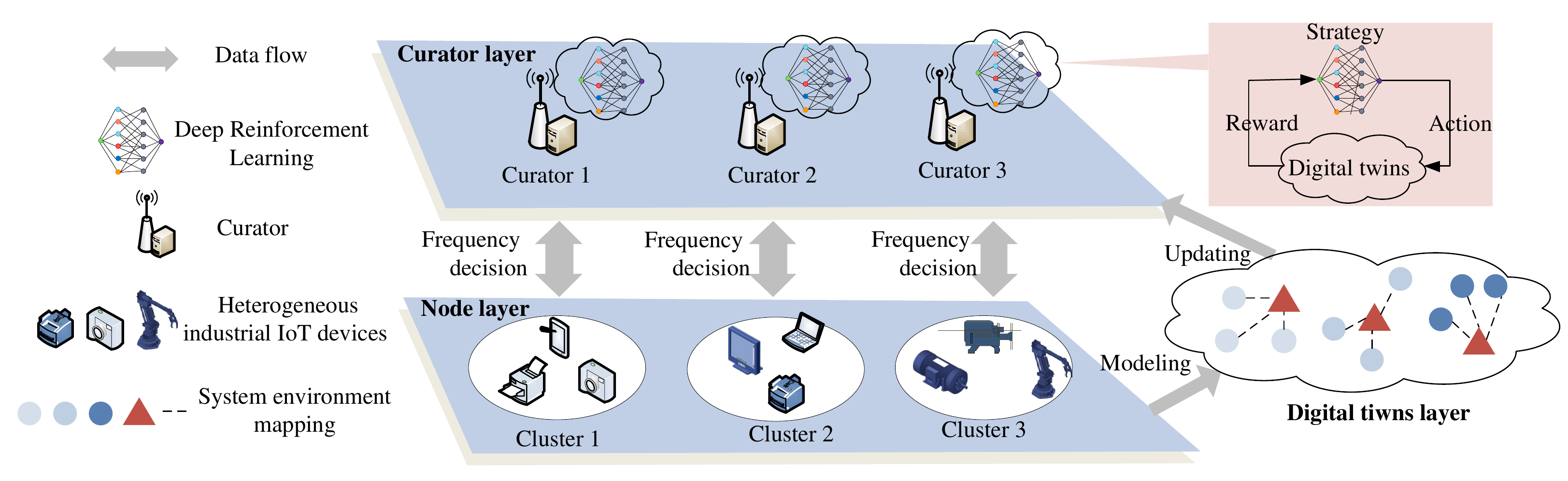}
\caption{DTs  for federated learning in a heterogeneous Industrial IoT scenario.}
\end{figure*}
\subsection{DTs for Industrial IoT}

As shown in Fig. 1, we introduce a three-tier heterogeneous network in Industrial IoT, consisting of industrial devices, servers, and DTs of industrial devices. Devices with limited communication and computing resources are connected to servers via wireless communication links. DTs are models that map the physical status of devices and update in real time.

The DT of an Industrial  device is established by the server to which it belongs, where the history and current behavior of the device are dynamically presented in a digital form by collecting and processing the existing key physical state of the device.  Within time $t$, the DT  of the training node $i$  $DT_i(t)$ can be expressed as 
\begin{equation}\label{1}
DT_i(t) =  \{F(w_i^t), f_i(t),  E_i(t)\},
\end{equation}
where $w_i^t$ is the current trained parameter of the node $i$, $F(w_i^t)$ represents the  current training state of the  node $i$, $f_i(t)$ is the current computational capability of the  node $i$, and $ E_i(t)$ indicates the energy consumption.

Note that  there is a deviation between the mapped value  of DT and the actual value. We use the CPU frequency deviation $\hat{f}_i(t)$ to represent the deviation between the actual value of the device and its DT mapping value.
Therefore, the DT model after calibration  can be expressed as
\begin{equation}
  \hat{DT}_i(t) =  \{F(w_i^t), f_i(t)+ \hat{f}_i(t),  E_i(t)\}.
\end{equation}
This model can receive the physical state data of the device and perform self-calibration according to the empirical deviation value, while maintaining consistency with the device and feeding back information in real time, thereby achieving dynamic optimization of the physical world.

\subsection{Federated learning in Industrial IoT}
In  Industrial IoT, industrial devices (excavators, sensors, monitors, etc.) need collaboratively complete production tasks based on federated learning \cite{27}.  As shown in Fig. 1, an excavator with sensors collects a large amount of production data. 
Through collaboration of curators to perform federated learning and intelligent analysis, better decisions can be made for quality control and predictive maintenance.

In Industrial IoT, when resource-constrained industrial devices (excavators, sensors, monitors, industrial robots, etc.) need complete manufacturing tasks, such as defective detection, they typically train a global model collaboratively based on federated learning using the their own data set. For example, as shown in Fig. 1, a machinery device collects a large amount of production data in a real-time monitoring environment and trains its detective model. Then each device would upload their local model (may be different from each other due to heterogeneous data set) to the curator (typically an edge server), which completes the aggregation and update of the model parameters, and returns the updated parameters to each machinery device. Each device repeats the next iteration until the entire training process converges. Through collaboration of machinery devices based on federated learning, intelligent decisions can be made for quality control and predictive maintenance.

The first step of federated learning is task initialization, in which the curator broadcasts the task and the initialized global model $w_0$.
Next, after receiving $w_0$, the training node $i$ uses its  data $D_i$ to update the local model parameters $w^t_i$ to find the optimal parameter that minimizes the loss function
\begin{equation}
F(w_i^t)=\frac{1}{D_i}  \sum_{x_j,y_j \in D_i} f(w^t_i,x_i,y_i),
\end{equation}
where  $t$ denotes the current local iteration index, $f(w^t_i)$ quantifies the difference between estimated and true values for instances of running data $D_i$, and $\{x_i, y_i\}$ is the samples of training data. 
 After $T$ rounds of local training, the node sends the updated local model parameters to the curator, where $T$ is a preset  frequency.
 Then, the curator is responsible for the global model aggregation to obtain the parameters of the $k$-th aggregation $w_k$ according to the preset aggregation strategy (see Subsection III-C). The loss value after the k-th global
aggregation is $ F(w_k)$.
    After global aggregation, the curator broadcasts the updated global model parameter  back to the training node. Local model training and global model aggregation  need to be repeated until the global loss function converges or reaches the preset accuracy.

\subsection{Trust-based Aggregation in DT-driven  Industrial IoT}

In federated learning in Industrial IoT, to improve the learning accuracy and resist malicious attacks, the parameters uploaded by nodes with high reputation should have greater weight in the aggregation. Unlike the traditional reputation model that only considers security threats, we comprehensively consider the effects of DT deviation, learning quality, and malicious data on learning.

Note that DTs have inevitable deviations in the mapping of node states in terms of CPU frequency $\hat{f}_i$,  and the mapping deviations for different nodes are different. The greater the DT error, the lower the curator’s belief for the node is. Therefore, the parameters uploaded by nodes with low mapping deviation should occupy more weight in the aggregation. In addition, in a \emph{Byzantine attack}, a malicious client user may provide the curator with incorrect or low-quality updates instead of an effective update. 
Therefore, we introduce learning quality and interaction records that count for malicious updates to weaken the threat of malicious data. Based on the subjective logic model, the  belief of the curator $j$ for the node $i$ in the time slot $t$ can be expressed as
 \begin{equation}
\label{eq6}
b_{i\rightarrow j}^t = \frac{(1 - u_{i \rightarrow j}^t)q_{i \rightarrow j}^t}{\hat{f}_i(t)} \cdot \frac{\alpha_i^t}{\alpha_i^t + \beta_i^t},
\end{equation}
where $\hat{f}_i(t)$ indicates the DT deviation of the curator $j$ to the node $i$, 
$\alpha_i^t$ is the number of positive interactions, and $\beta_i^t$ is the number of malicious actions such as uploading lazy data, 
$q_{i \rightarrow j}^t = \frac{w_i^t-\bar{w}}{\sum_{i=1}^{n}w_i^t-\bar{w}}$ denotes quality of learning based on the honesty of most training devices. The difference between the parameters uploaded by a node and the average value of the parameters uploaded by all nodes positively affects the learning quality of a client. From (4), we can see that the higher the learning quality of a client, the higher its reputation value will be. Moreover, the curator will also choose the node with higher reputation value in order to improve the learning accuracy and resist malicious attacks.
Specifically,  the curator uses $FoolsGold$ scheme that identifies unreliable nodes according to the gradient update diversity of local model updates in non-IID federated learning [12].  The reputation value of the curator $i$ for the node $j$ is expressed as
\begin{equation}
  T_{i\rightarrow j} = \sum_{t=1}^{T} b_{i\rightarrow j}^t + \iota u_{i\rightarrow j}^t,
\end{equation}
where $\iota \in [0,1]$ is a coefficient indicating the degree of uncertainty affecting the reputation and  $u_{i \rightarrow j}^t$ represents the failure probability of packet transmission. In the global aggregation,  the curator retrieves the updated reputation value and aggregates the local model  $w_i^k$ of the participating nodes into a weighted local model, i.e.,
\begin{equation}
  w_k=\frac{\sum_{i=1}^{N_d}\sum_{t=1}^{T} T_{i\rightarrow j} w_i^t}{\sum_{i=1}^{N_d} T_{i\rightarrow j}},
\end{equation}
where $w_k$ represents the global parameter after the $k$-th global
aggregation, $N_d$  is the number of training device.
Through such a trusted-based aggregation, the deviation of DT is considered and the security threats caused by malicious participants are effectively resisted, which can enhance the robustness of the framework while increasing the learning convergence rate.

\subsection{Energy Consumption   Model in Federated Learning}

 In  federated learning, the energy consumption of a training node is composed of the computational energy consumed by local training and the communication energy consumed by global aggregation \cite{28}. The  computational energy consumed by training node $i$ to perform one training is expressed as
\begin{equation}\label{eq:w2}
  E^{cmp}=n_{cmp}F/f_i,
\end{equation}
where $F$ is defined as the CPU frequency required for one training and  $n_{cmp}$ is the normalization factor of the consumed computing resources. To formulate the communication energy consumed, we use $M$ to represent the number of bits of the neural network model. 
In order to transmit the parameters to the curator, all training nodes share $F$ uplink sub-channels based on orthogonal frequency division multiple access (OFDMA), which is expressed as a set $\mathcal{C} = \{1,2, ..., C\}$. The model parameters mentioned here refer to the gradients. Therefore, there is no co-channel interference between the training nodes. The curator will first collect the model parameters obtained after the local training of all nodes is completed, then aggregate these parameters, and finally send the aggregated parameters to all nodes through all sub-channels. The communication resources consumed by training node i to perform an aggregation are expressed as
\begin{equation}\label{eq:w1}
  E^{com}=\frac{n_{com}M}{\sum_{c=1}^{C}l_{i,c}W log_2(1+\frac{p_{i,c}h_{i,c}}{I})},
\end{equation}
where $l_{i, f}$ represents the time fraction allocated by the training node $i$ on sub-channel $f$, $W$ denotes the sub-channel bandwidth, $p_{i,f}$ is the uplink transmission power of the training node $i$ on sub-channel $n$, $h$ is the uplink channel power gain between the training node $i$ and the curator,  $I$ is the noise power and  $n_{com}$ is the normalization factor of the consumed communication resources.

\subsection{Problem Formulation}
The objective of this paper is to determine the best tradeoff between local update and global parameter aggregation  in a time-varying communication environment with a given resource budget to minimize the loss function.
The aggregation frequency problem can be formulated as
 \begin{align}
\mathbf{P1:}& \mathop{arg \min}_{ \{a_0, a_1, ... , a_k\} } F(w_k) \label{YY}\\
& s.t.  \,  \sum_{i=1}^{n}a_i  E^{cmp}  + k E^{com} \leq \beta R_m \tag{\ref{YY}{a}} \label{YYa}
\end{align}
where  $w_k$ represents the global parameter  after the $k$-th global aggregation,  $F(w_k)$ is the loss value after the $k$-th global aggregation, $\{a_0, a_1, ... , a_k\}$ is a set of strategies for the frequency of local updates,  $a_i$ indicates the number of local updates required for the $i$-th global update.
Constraint (\ref{YYa}) represents the given budget on available resources, and $\beta$ represents the upper limit of the resource consumption rate in the entire learning process.  In Eqn. (9) and Eqn. (9a), the loss value $F(w_k)$ and the  computational energy consumption $E^{cmp}$ include the training state $F(w_i^t)$ and  computational capability $f(i)$, respectively, which is estimated by DTs to enable the curator to grasp the key status of the entire federated learning. The deviation in computational energy consumption $ E^{cmp}$ due to the mapping deviation of DT in the  computational
capability of the node is calibrated by  trust-based aggregation.

\section{DEEP REINFORCEMENT LEARNING FOR Aggregation frequency based on Digital Twin}
\subsection{Problem Simplification}

Due to the long-term resource budget constraints, it is difficult to solve \textbf{P1}. If the  energy consumption at the current aggregation is too high, it will lead to energy shortage in the future. In addition, \textbf{P1} is a nonlinear programming problem, and the complexity increases exponentially with the increase of federated learning rounds. Therefore, it is necessary to simplify \textbf{P1} and the long-term resource budget constraints.

The effect of training after $k$ global aggregations can be written as
\begin{equation}\label{eq8}
F(w_0)-F(w_k)=\sum_{i=1}^{k}[F(w_{i-1})-F(w_i)].
\end{equation}
For optimal training results, i.e., 
\begin{equation}
\min [F(w_k)-F(w_0)]=\max \sum_{i=1}^{k}[F(w_{i-1})-F(w_i)].
\end{equation}
 Based on the \emph{Lyapunov optimization}, a dynamic resource deficit queue is established to simplify \textbf{P1} by dividing the long-term resource budget into available resource budget for each time slot.  Lyapunov optimization has been widely used in control theory, which can make the system stable in different forms. Firstly, we construct a virtual queue that satisfies the constraints of \textbf{P1} for Lyapunov optimization, and then we define the Lyapunov function to represent the backlog of all virtual queues in time slot t. Secondly, we design the difference of the Lyapunov function between two subsequent time slots. Finally, by minimizing the Lyapunov drift of each time slot, and always pushing the backlog to a lower congestion state, we can intuitively maintain the stability of the queue. We define the length of the under-resource queue as the difference between used resources and available resources \cite{28}. The limit of the total resource is $R_m$, and the resource available in the $k$-th aggregation is $\beta R_m/k$.  The evolution of the resource deficit queue is as follows
\begin{equation}
  Q(i+1)=\max \{Q(i)+(a_i E^{cmp}+ E^{com})-\beta R_m/k,0\},
\end{equation}
where $(a_iE^{cmp}+ E^{com})-\beta R_m/k$ is the deviation of resources in the $k$-th aggregation.  According to the above Eqn. (8) and Eqn. (11), the original problem \textbf{P1} can be transformed into
 \begin{align}
 \mathbf{P2}:   \mathop{arg \max}_{ \{a_0, a_1, ... , a_k\} } \sum_{i=1}^{k}& [v(F(w_{i-1})-F(w_i)) \notag\\
 &-Q(i)( a_i E^{cmp} + E^{com})] \label{w1}\\
  s.t. \, \sum_{i=1}^{n}a_i  E^{cmp}&  + k E^{com} \leq \beta R_m \tag{\ref{w1}{a}}\label{Ya}
 \end{align}
where $v$ and $Q(i)$ are positive control parameters that dynamically balance training performance and resource consumption. It is noted that the accuracy of federated learning can be easily improved at the beginning of the training, while it is costly to improve the accuracy at the later stage. Therefore, $v$ increases with the increase of training rounds and is motivated towards the goal of maximizing the ultimate benefits.

\subsection{Markov Decision Process (MDP) Model}
We use deep reinforcement learning (DRL) to solve the frequency problem of local updates, DT learns models by interacting with the environment, without requiring pre-training data and model assumptions. 
 We formulate the  optimization problem as a MDP denoted by $\mathcal{M}$, which includes the system state $\mathcal{S}(t)$, action space $\mathcal{A}(t)$, policy $\mathcal{P}$, reward function $\mathcal{R}$  and next state $\mathcal{S}(t+1)$. The detailed description of parameters is as follows:

 \begin{itemize}
   \item  \textbf{System State} The system state describes the characteristics and training state of each node, including the current training state of all nodes $\varsigma(t)$, the current state of the resource deficit queue $Q(i)$, and the average value output from the single hidden layer with 200 neurons in the neural network of each node $\tau(t)$. The current training state is directly related to the aggregation strategy, it represents the training degree of local nodes. The average value output reflects the categorical characteristics of the sample to a certain extent, which can effectively avoid multiple training on invalid samples and improve training efficiency, i.e.,
       \begin{equation}
         \mathcal{S}(t)=\{\varsigma(t), \tau(t), Q(i), \mathcal{A}(t-1)\}.
       \end{equation}
   \item \textbf{Action Space} The action is defined by a vector $\mathcal{A}(t)=\{a_1^t, a_2^t, ... , a_n^t \}$ where $a_i^t$ indicates the number of local updates, which need to be discretized. For simplicity, we use $a_i$ to denote $a_i^t$ because our subsequent statements are based on a specific time $t$.
   \item \textbf{Policy} Policy $\mathcal{P}$ is the mapping of state space to action space, i.e., $\mathcal{A}(t) = \mathcal{P}\{\mathcal{S}(t)\}$, to verify whether the local model update provided by the node is trusted.
     \item  \textbf{Reward Function} It is noted that our goal is to determine the best tradeoff between local update and global parameter aggregation to minimize the loss function, the reward function is naturally related to the decline of the overall loss function and the state of the resource loss queue. The action is assessed by
     \begin{equation}\label{reweq}
       \mathcal{R}=[vF(w_{i-1})-F(w_i)]-Q(i)( a_i E^{cmp} +  E^{com} ).
     \end{equation} 
     It can be found that the consumed computational energy is a factor affecting the reward function, which is also affected by the DT deviation. Due to the existence of DT deviation, the actual reward value may deviate from the expected reward, resulting in unstable and unreliable decision-making, which is very likely to invalidate the learning process. In order to solve this issue, we add the weight of node reputation value in (4). After that, CPU frequency error can be corrected by the corresponding weighting factor, which reduces the influence of DT deviation on model learning.

     \item \textbf{Next State}  The current state $\mathcal{S}(t)$ is provided by the DT real-time mapping, and the next state $\mathcal{S}(t+1)$ is the prediction obtained by the DT running DQN without actually running in the physical world, which can be expressed as $\mathcal{S}(t+1) = \mathcal{S}(t)+ \mathcal{P}(\mathcal{S}(t))$. 
 \end{itemize}

In fact, the reward function here is just immediate reward. The whole reward function includes the expectation of immediate reward and the expectation of subsequent reward. The subsequent reward expectation is usually multiplied by an attenuation coefficient $\gamma$ within 0 and 1 to ensure convergence. When the current state, action, and new state are confirmed, the expectation of immediate reward can be determined. Moreover,  $\lambda$ is pre-determined in our paper, therefore, when the agent adopts a completely greedy strategy, the state value obtained by making a choice in the current state is can be also determined. Meanwhile, in our paper, we adopt strategy improvement plan, which guarantees the incremental nature of the function value. Particularly, we calculate all corresponding $q_{\pi}(s,a)$  to get $a$, and compare the maximum value, which inevitably leads to the result $v_{\pi}(s)\leq  q_{\pi}(s,a)$ where $a = \arg max q_{\pi}(s,a)$.
	
We first leverage a random action selection strategy during \emph{DQN} training so that the reward value corresponding to the state can be recorded. Then we slowly increase the greedy coefficient until it reaches 1. Under the completely greedy action selection strategy, if we set $\pi(S_t)$ as the strategy composed of all state actions under iteration t, and $\pi (S_{t+1})$ as the strategy in the case of iteration $t+1$, if $\pi(S_t) = \pi (S_{t+1}) $, it can be indicated that the strategy is convergent. According to the previous analysis, function non-convergence will inevitably lead to oscillation, which also leads to continuous improvement of strategic value. However, the strategy space is limited, the fixed point of the strategy value is also limited. One of the fixed points must be the largest, which is the optimal fixed point. Actually in our simulation diagram, the fixed point with the largest value exists, which is contradictory to continuous improvement. That is, our strategy is convergent.

In the proposed scheme, the state space includes the initial resource value of each node and the corresponding model loss value. The action space includes the number of training sessions for each node, determined by the curator. The initial state is randomly chosen. The previous state and the model loss value can jointly determine the number of training required by the node in the current state, that is, the action in the current state. Then the action will cause the node resource value and the model loss value to change, that is, the next state.

\subsection{DQN-based Optimization Algorithm for Aggregation Frequency}

\begin{algorithm}[t]
			\caption{Adaptive calibration of the global aggregation frequency.}
			\label{alg:ADRA}
			\LinesNumbered
			\KwIn{ eval\_net $O$, target\_net $O'$, update frequency $F_u$ of target\_net parameters, greed coefficient $\epsilon$, greed coefficient growth rate $r$;}
			\KwOut{The parameters of the trained DQNs;}
			Randomly initialize the parameters of evaluation eval\_net $O$ and target\_net $O'$ \;
			\For{each episode}{
				 Initialize the parameters in environment setup \;
				
               \For{each time slot t}{
                select $a_i^t=\mathop{\max}_{\mathcal{A}}O(\mathcal{S}(t),\mathcal{A};w_i)$ with probability $\epsilon$ and select a random action $a_i^t$ with probability $1- \epsilon$\;
                Perform federated learning training\;
                Calculate immediate reward $\mathcal{R}$ with Eqn. (15) and update the system state $\mathcal{S}(t+1)$\;
                Store the experience tuples $(\mathcal{S},\mathcal{A},\mathcal{R},\mathcal{S}(t+1))$\;

                \If {the experience relay is full}{
                \If{$t \% F_u=1$}{
                Update the target\_net parameters}
                 Learning all samples from the experience relay\;
         Calculate   q-eval value by  eval\_net\;
         Calculate  q-target value according to Eqn. (17)\;
            Perform a gradient descent step according to Eqn. (18)\;
                }{ }
                             }
                }
				\Return{The parameters of the trained DQNs.}
		\end{algorithm}

To solve the MDP problem, we use \emph{DQN}-based optimization algorithm.
As shown in Fig. 1,  DTs map all aspects of the physical objects of the Industrial IoT environment to the virtual space in real time, forming a digital mirror image. At the same time, the DRL agent interacts with the DTs of the devices to learn the global aggregation frequency decision.  The Federated learning module makes frequency decisions based on the trained model and the DT status of the training nodes. Through DTs, the agent achieves the same training effect as the real environment at a lower cost.

\subsubsection{Training Step}

 When using \emph{DQN} to achieve adaptive calibration of the global aggregation frequency, we first assign initial training samples to the training nodes and  set initial parameters for the target\_net and the eval\_net to maintain their consistency.  The state array consists of the initial resource value and the corresponding loss value obtained by training each node.  In each iteration,  the state of the experience relay needs to be judged. 
 If it is full, the action is selected with probability  according to the $\epsilon-$greedy strategy, otherwise the action is randomly selected with probability $\epsilon$.

 After the action is selected,  the reward is calculated according to Eqn.  (\ref{reweq}) and the system status is updated. Next,  the  current state, selected actions, rewards, and the next state are recorded in the experience relay. Then we sample from the experience relay to train target\_net, which randomly disrupts the correlation between the states by randomly sampling several samples in the experience replay as a batch.  By extracting the state, the eval\_net parameters are updated according to the loss function as follows:
\begin{equation}
  F(w_i)=\mathbb{E}_{\mathcal{S},\mathcal{A} }[y_i-O(\mathcal{S},\mathcal{A};w_i)^2],
\end{equation}
where $O (\mathcal{S},\mathcal{A}, w_i)$ represents the output of the current network eval\_net, and $y_i$ is the q-target value calculated according to the parameters of the target\_net, which is independent of the parameters in the current network structure.
The target\_net is used to calculate the q-target value  according to the following formula:
\begin{equation}
  y_i=\mathbb{E}_{\mathcal{S},\mathcal{A}}[r+\gamma \max_{\mathcal{A}'}O(\mathcal{S}', \mathcal{A}';w_{i-1})| \mathcal{S},\mathcal{A}],
\end{equation}
where  $\{\mathcal{S}',\mathcal{A}'\}$  is the sample from  the experience relay, $O (\mathcal{S}', \mathcal{A}', w_{i-1})$ represents the output of target\_net.
 In this way, the entire objective function can be optimized by the stochastic gradient descent method:
\begin{equation}
\begin{aligned}
\triangledown_{w_i}F(w_i)=&\mathbb{E}_{\mathcal{S},\mathcal{A} }[(r+ \gamma \max_{\mathcal{A}'}O(\mathcal{S}',\mathcal{A}';w_{i-1})\\
& -O(\mathcal{S},\mathcal{A};w_i))\triangledown_{w_i}O(\mathcal{S},\mathcal{A};w_i)].
\end{aligned}
\end{equation}

After a certain number of iterations, eval\_net parameters need to be copied to target\_net. Namely, the updates of loss and target\_net are performed in time intervals and experience replay is updated in real time. Repeat the above steps until the loss value reaches the preset value.  The complete frequency algorithm for global update of a single cluster is presented in \textbf{Algorithm 1}.

\subsubsection{Running Step}
 After training, the proposed aggregation frequency decision agent is deployed on the curators to make the optimal decision according to the users’ DTs. In this way, changes in the network topology and digital twin status can be quickly captured, and a new model can be quickly established based on the training results accumulated in the previous period.

\subsection{ DQN-based Asynchronous Federated Learning }
\begin{algorithm}[t]
			\caption{Intra-cluster aggregation frequency decision.}
			\label{alg:ADRA}
			\LinesNumbered
			\KwIn{ The trained DQNs;}
            \KwOut{Local aggregation parameters;}
       For each cluster, run the DQN trained by \textbf{Algorithm 1} on the curator\;
			
				\For{ each global aggregation}{
               \For{ aggregation frequency $a_i$ given by DQN}{
                \If {$a_if_i$ $>$ $\alpha T_m$}{
                $a_i = \lfloor \alpha T_m / f_i \rfloor $
                }
                Perform $a_i$ local trainings\;
              }
              }
		\end{algorithm}
In an Industrial IoT scenario where devices are highly heterogeneous in terms of available data sizes and computing resources,  single-round training speed is likely to be limited by the slowest nodes, i.e., the straggle effect, and synchronous training  is  not applicable. Therefore, we propose an asynchronous federated learning framework. The basic idea is to classify nodes with different computing power by clustering  and configuring corresponding curators for each cluster to enable each cluster to train autonomously with different local aggregation frequency. Within a cluster, the aggregation frequency is obtained through the DQN-based adaptive frequency calibration algorithm mentioned in Section IV-C. The specific asynchronous federated learning process is as follows.

  \textbf{Step 1: Node clustering.} 
 We first use clustering algorithm \emph{K-means} to classify nodes according to data size and computing power [12] and assign the corresponding curators to form a local training cluster.
 
 In this way, the execution time of each node in the same cluster of the local model is similar and does not drag each other.

 \textbf{Step 2: Aggregation  frequency decision.}
    Each cluster runs   Algorithm 2 separately to obtain the corresponding global aggregation frequency. To match the frequency with the computing power of the node, we use the minimum time $T_m$ required for local update  in the current round as a benchmark and specify that the local training time of other clusters cannot exceed $\alpha T_m$ where $\alpha$ is the tolerance factor between $0$ and $1$.  The tolerance factor $\alpha$ increases as the round of global aggregation increases, and the intuition behind this approach is that the effect of global aggregation on learning efficiency decays as the round increases. 

\textbf{Step 3: Local aggregation.} 
      After completing the local training according to the frequency given by \emph{DQN}, the curators of each cluster use the trust weighted aggregation strategy to locally aggregate the parameters uploaded by the nodes. Specifically, the curator needs to retrieve the updated credit value and measure the importance of different node upload parameters according to Eqn. (7). Parameters uploaded by nodes with low mapping deviation and high learning quality take up more weight in local aggregation, which improves the accuracy and convergence efficiency of the model.

 \textbf{Step 4: Global aggregation.} Finally,  the time-weighted aggregation is used to aggregate global parameters.  
In order to distinguish the contribution of each local model to the aggregation operation  based on time effect and improve the effectiveness of the aggregation operation, once the time for global aggregation arrives, the curators upload the parameter $w^t_j$ with the time version information, and the selected curator performs the global aggregation as
\begin{equation}\label{eqtamp}
w_{k+1} \leftarrow \sum_{j=1}^{N_c}(e/2)^{-(t-timestamp^k)}w^t_j,
\end{equation}
where  $N_c$  is the number of curator, $w^t_j$ is the aggregated parameter of cluster $j$, $e$ is the natural logarithm used to describe the time effect,   and $timestamp^k$ is the timestamp corresponding to the latest parameter of $w^k_j$, i.e., the number of rounds.

Through the proposed   heterogeneous framework equipped with trust mechanism, the straggler effect is eliminated and malicious node attacks are effectively evaded, naturally improving the convergence rate and learning quality. 
Traditional weighted average strategy adopting fixed frequencies for local training has its defects. On the one hand, it ignores the impact of differences between samples, which are prone to problems such as overfitting. On the other hand, the decision-making process often only considers the accuracy of training, and cannot optimize the problem from multiple dimensions. Although the \emph{DQN}-based approach requires a large number of samples for training, it can solve the problem including sample distribution, resource consumption and so on, which improves the effect of training while maintaining realistic meaning. Moreover, the \emph{DQN} training is implemented only when the network topology or channel condition changes, thus the time consumption of \emph{DQN} training averaged over the whole process is acceptable. 


\section{Numerical Results}

We assume that devices in the Industrial IoT need to recognize each other and cooperatively perform production tasks based on federated learning such as defective product detection. Based on the publicly available large image data set MNIST, we apply the proposed scheme to real object classification tasks.
We implement asynchronous federated learning and \emph{DQN} in PyTorch. \emph{DQN} is initialized by two identical neural networks, where the size of each network is $48\times200\times10$  consisting of three fully connected layers deployed in sequence. As there is an error in the mapping of DT to the real CPU frequency, we make the DT error subject to a uniform distribution between 0 and 0.2. Moreover, we assume the noise distribution all obey the Poisson distribution, and the mean value of the noise influence is 0.1, 0.3 and 0.5 dB under good channel state, medium channel state and bad channel state respectively. To  illustrate the performance of the proposed scheme, we choose the fixed aggregation frequency scheme as the
benchmark scheme.


\begin{figure}[t]
\centering
\includegraphics[scale=0.36]{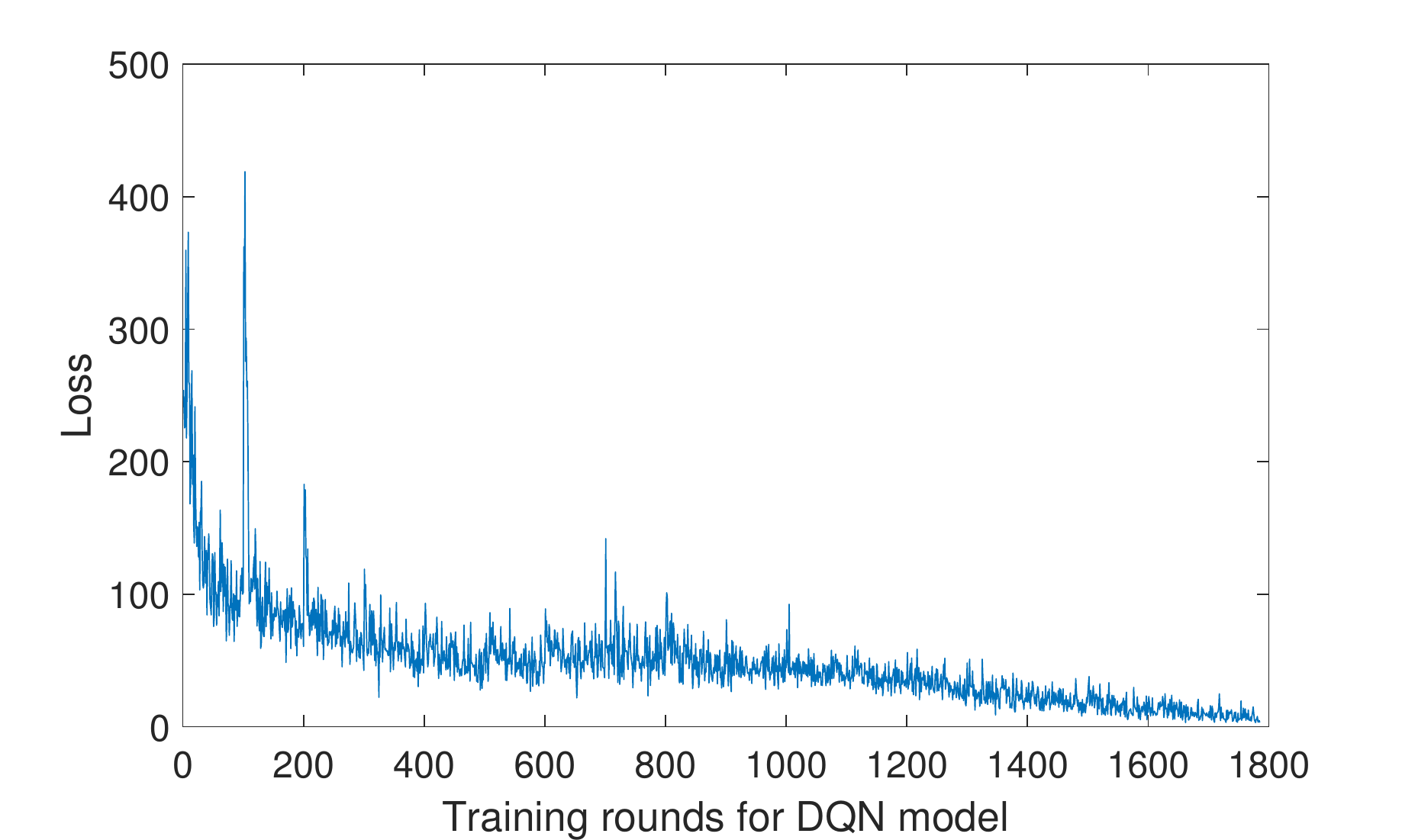}
\caption{The convergence performance of DQN.}
\end{figure}

\begin{figure}[t]
\centering
\includegraphics[scale=0.36]{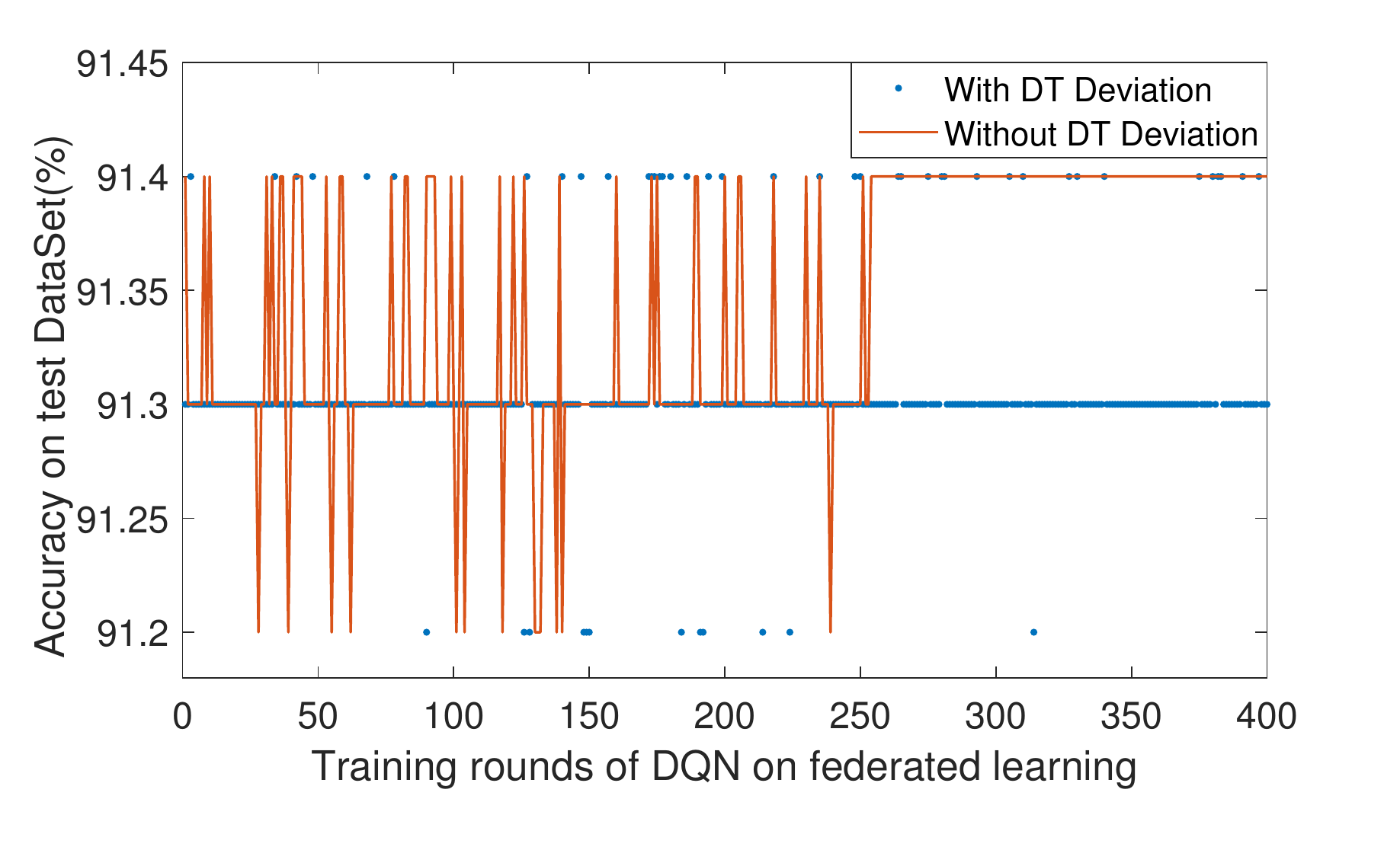}
\caption{Comparison of DT deviation and without DT deviation.}
\label{f4}
\end{figure}

\begin{figure}[t]
\centering
\includegraphics[scale=0.36]{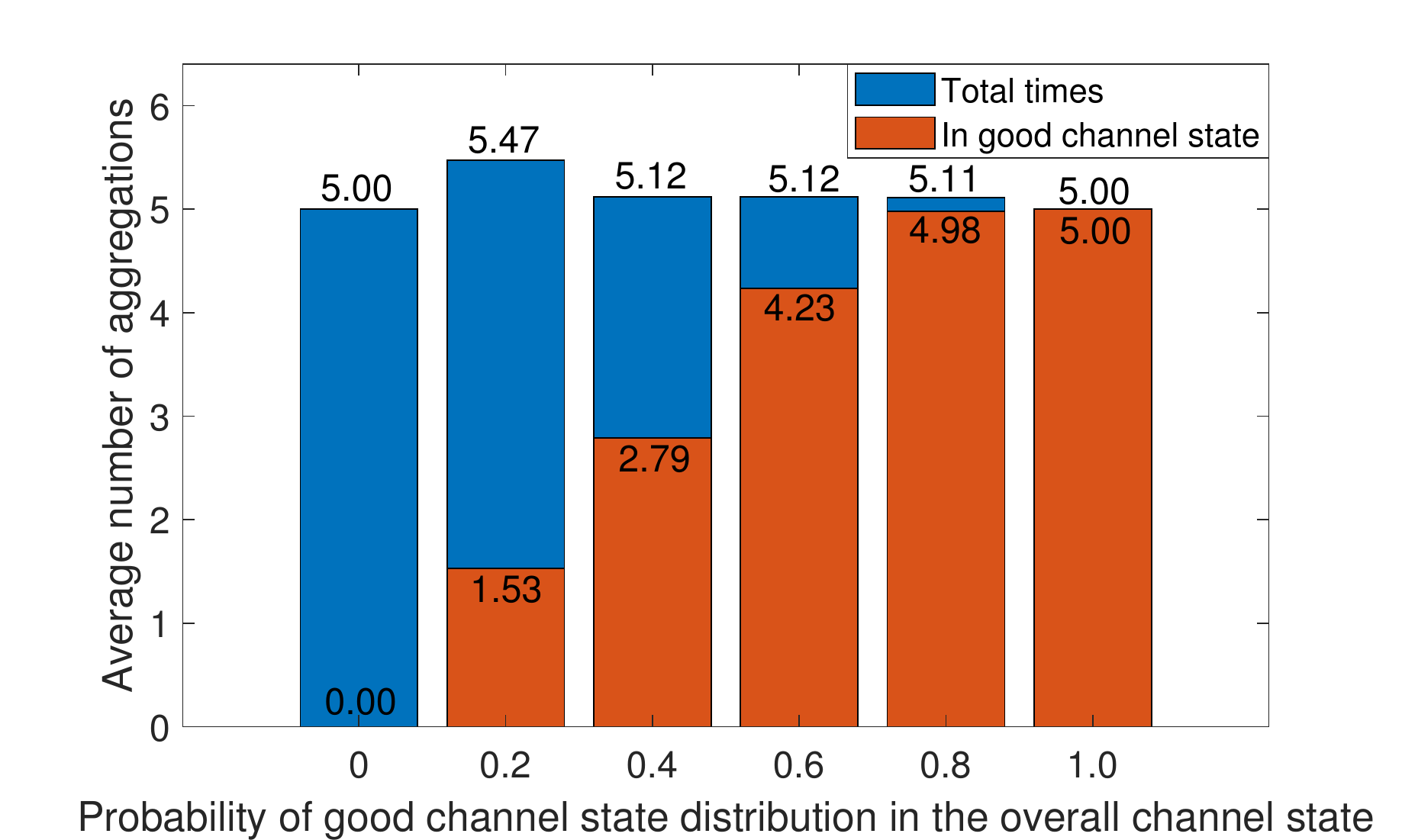}
\caption{The change of the average aggregation probability in a good channel state varies with the channel state.}
\label{f7}
\end{figure}

\begin{figure}[t]
\centering
\includegraphics[scale=0.36]{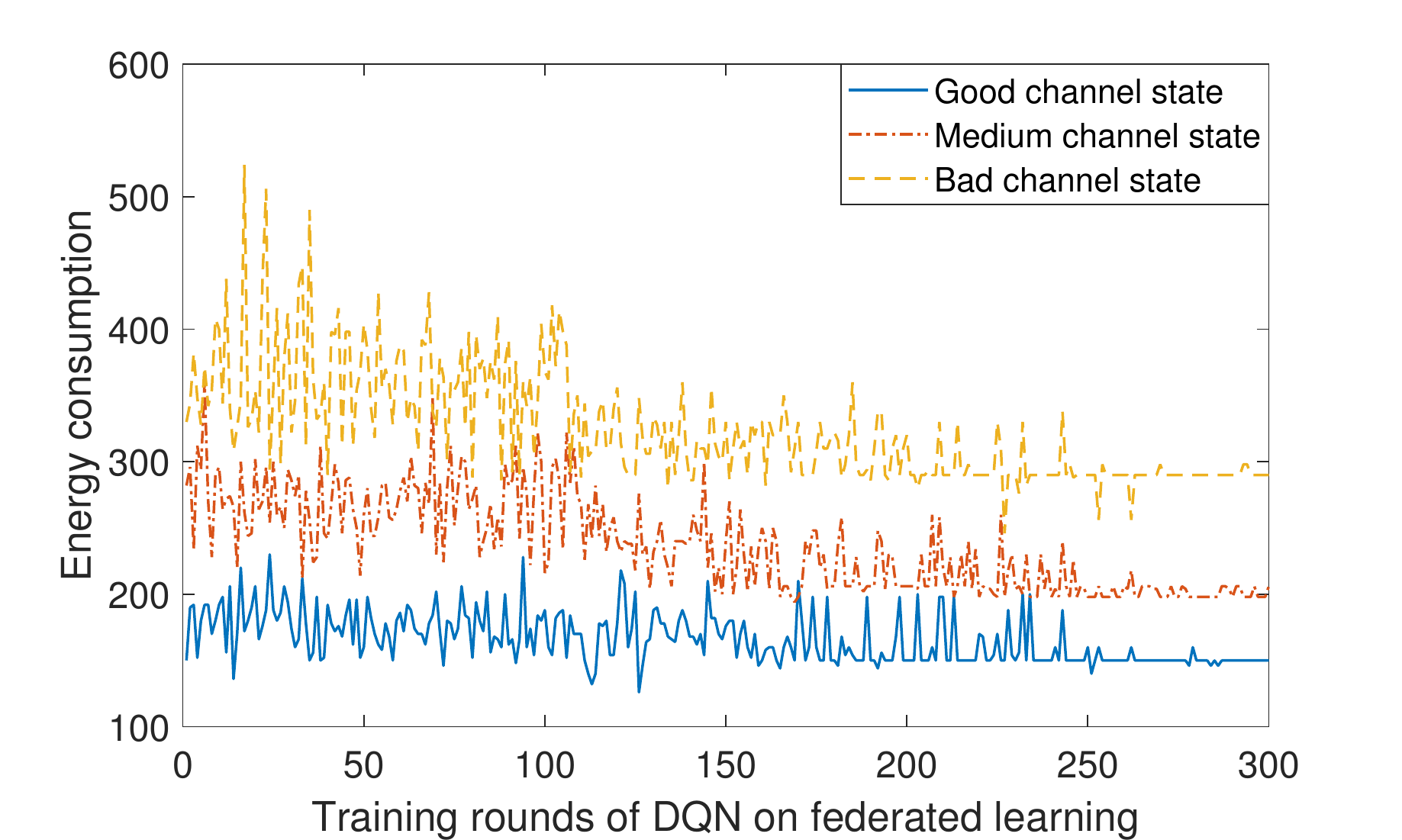}
\caption{Comparison of energy consumed by federated learning in different channel states. }
\label{f6}
\end{figure}

\begin{figure}[t]
\centering
\includegraphics[scale=0.36]{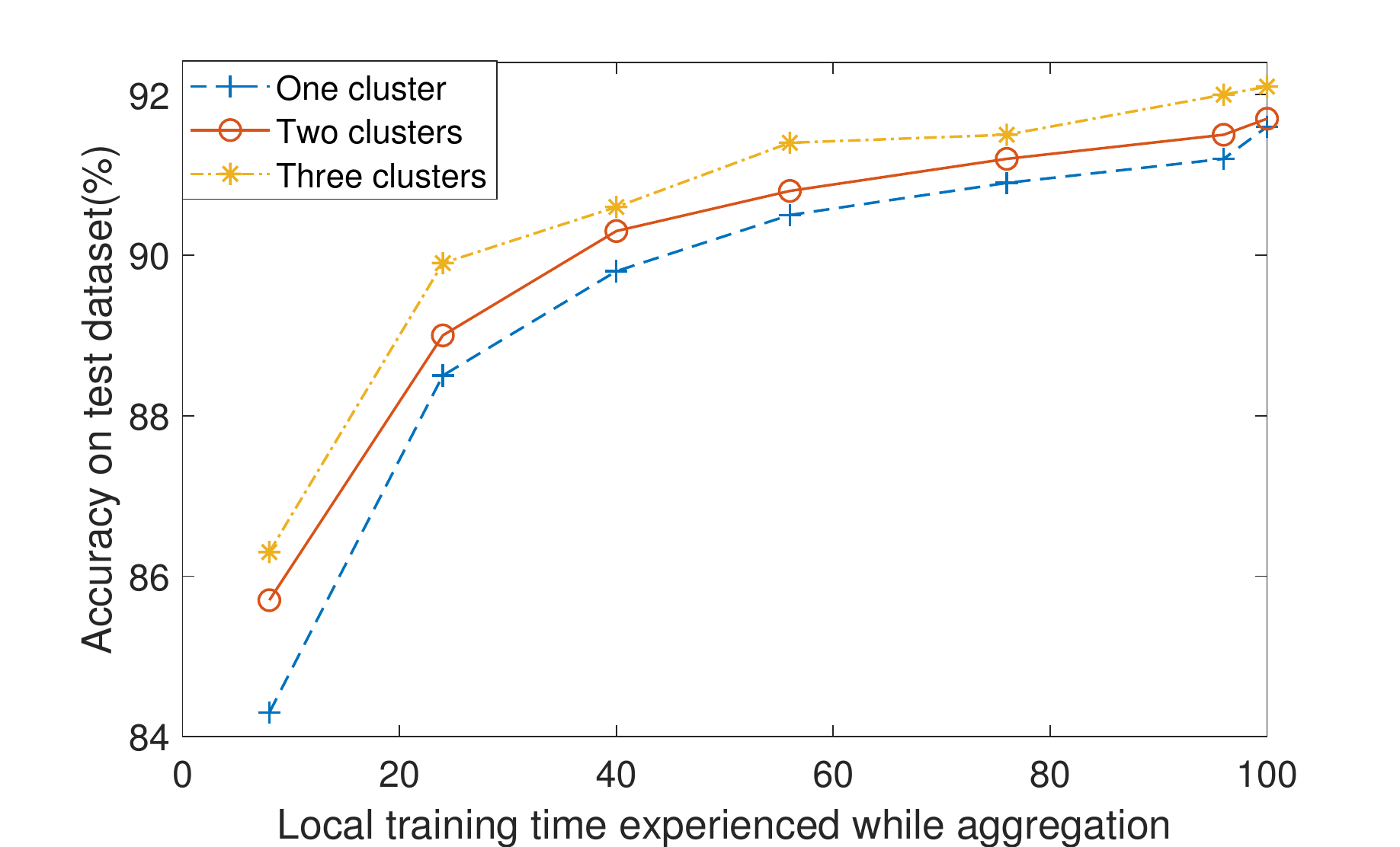}
\caption{Comparison of the accuracy achieved by federated learning under different clustering conditions.}
\label{f8}
\end{figure}

\begin{figure}[t]
\centering
\includegraphics[scale=0.36]{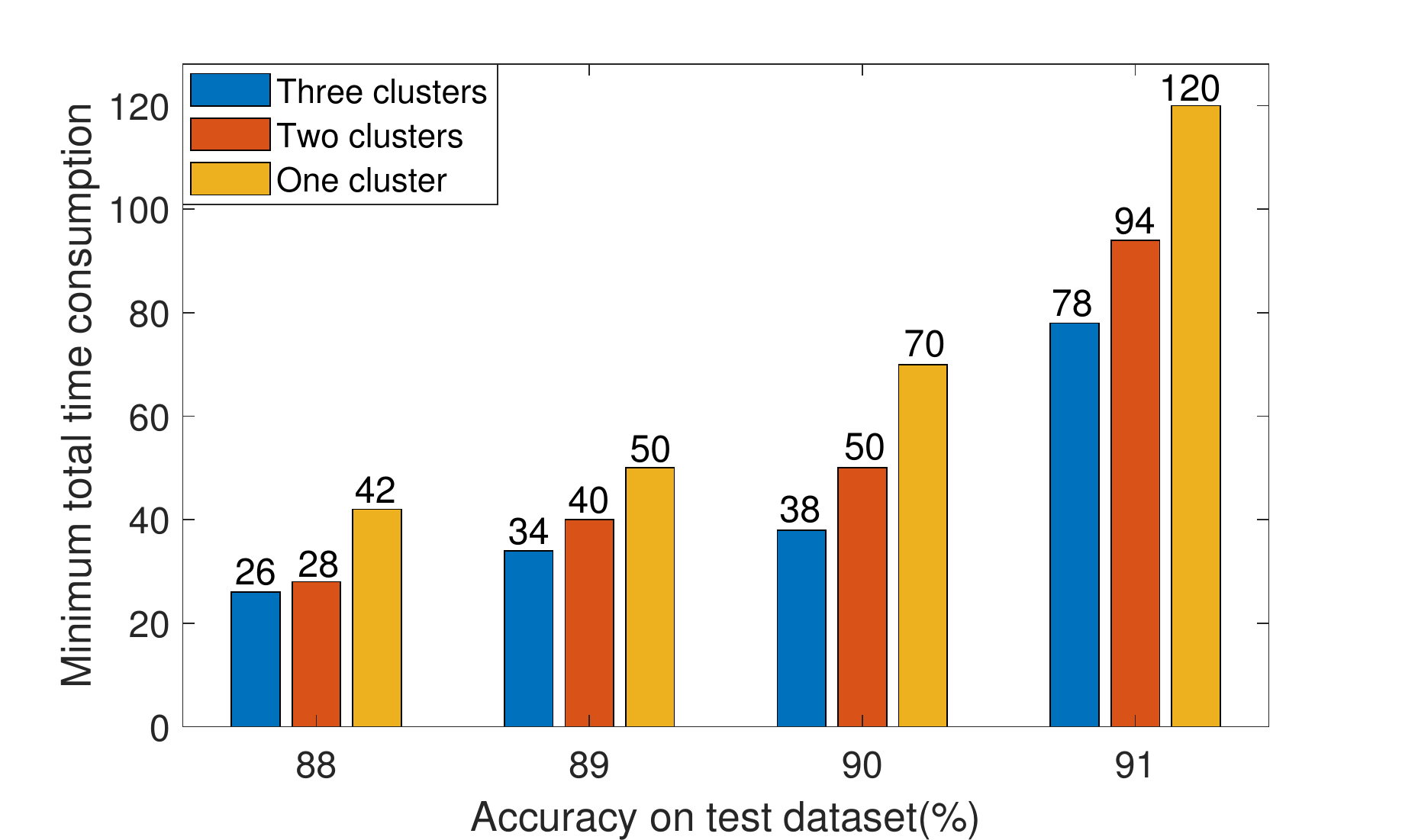}
\caption{Comparison of the time consumed by federated learning under different clustering conditions.}
\label{f9}
\end{figure}

\begin{figure}[t]
\centering
\includegraphics[scale=0.36]{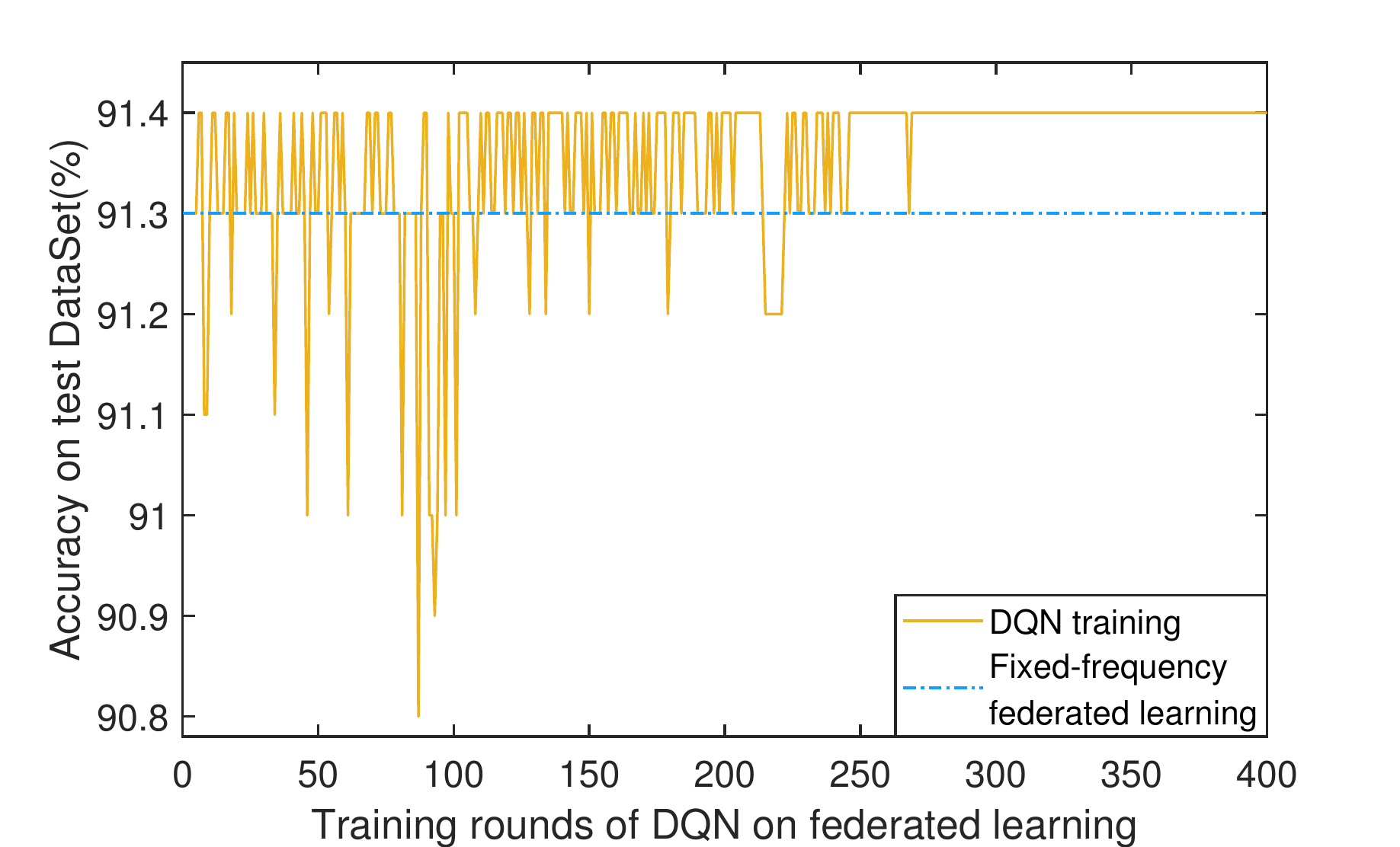}
\caption{Comparison of the accuracy achieved of federated learning between adaptive frequency and fixed frequency.}
\label{f10}
\end{figure}




Reinforcement learning is based on reward hypothesis that all goals can be described as maximizing the expected cumulative reward. The reward function is set to guide agent activities, and poorly designed rewards generally cause the network to not converge or the agent simply does not experience what it wants to learn. Fig. 2 depicts the trend of the loss value,  from which we can see that agent's loss has stabilized after about 1200 rounds of training and converged to a better result. Therefore,  the trained \emph{DQN} is suitable for heterogeneous scenarios and has good convergence performance.

 Fig. 3 compares the federated learning accuracy that can be achieved in the presence of DT deviation and after calibrating DT deviation.  The federated learning with  DT deviation calibrated by the trust weighted aggregation strategy can achieve higher accuracy than  the federated learning with the DT deviation, and the federated learning after calibration deviation is also superior when these two algorithms have not converged. In addition, we can observe that \emph{DQN} with DT deviation cannot converge.

%



Fig. \ref{f7}  shows  the total number of aggregations required to complete federated learning and the number of aggregations in a good channel state as the channel state changes. As the distribution of good channel states increases, the number of aggregations in good channel states increases. It is noted that almost all   aggregations are completed within 5 rounds  due to the fact that DQN learning finds that the benefits of less aggregation times are greater. Moreover, it can be seen that when the probability of being in a good channel equals 0.2, the total aggregation number is more than other cases. This is because when the channel state is varying and sometimes poor, the agent tends to wait until good channel state to make aggregation. With frequent bad channel state, as shown in the case of 0.2, the agent is prone to generate local optimal strategies, which leads to larger number of average aggregations, i.e. 5.47. There is an exception when the probability of good channel equals 0. Because when the channel state is constant, the agent won’t wait for the good channel state and will make aggregation anyway at higher communication cost, thus the average number of aggregations is also 5, the same as that of 1. In other cases, when the good channel ratio increases (higher than 0.2), it is easier for the agent to make aggregation adaptively at good channel state and make global optimal strategies (the average number of aggregations approaching 5). It shows that through continuous learning, \emph{DQN} intelligently avoids performing aggregation in poor channel conditions.

Fig. \ref{f6} compares the energy consumed by federated learning during \emph{DQN} training  in different channel states, in which the energy consumption includes computing resources during local training and communication resources during aggregation. It can be seen that energy consumption decreases with the improvement of channel quality, which is mainly due to the fact that aggregation consumes more communication resources when channel quality is poor. With the training of \emph{DQN}, the energy consumption in all three channel states decreases. This is because \emph{DQN} can adaptively calibrate aggregation timing so that federated learning selects local training instead of long-delay and high-energy-consuming aggregation when the channel quality is relatively poor.

Fig. \ref{f8} describes the variation of the accuracy achieved by federated learning in different clustering cases. It can be seen that the more clusters, the higher accuracy the training can achieve in the same time, which is because the clustering effectively uses the computing power of heterogeneous nodes through different local aggregation times. 
 Fig. \ref{f9} depicts the time required for federated learning to achieve preset  accuracy under different clustering situations. The training time required to achieve the same accuracy decreases as the number of clusters increases. Similar to Fig. \ref{f8}, this benefits from clustering effectively utilizing the computational power of heterogeneous nodes to make the local aggregation timing of different clusters different.
With the increase of the number of clusters, the straggler effect can be alleviated more effectively, which naturally shortens the time required for federated learning. In addition, when the preset accuracy reaches $90\%$ or higher, the same accuracy improvement takes more time.

Fig. \ref{f10} compares the accuracy  by DQN-based federated learning with that of fixed frequency federated learning. We can find from the training process  that $DQN$ learns and  then surpasses the accuracy value of fixed frequency. This is because the gain of the global aggregation to the federated learning accuracy is non-linear and the fixed frequency scheme may miss the best opportunity for aggregation. The accuracy of federated learning ultimately achieved by the proposed scheme is greater than that of the fixed frequency scheme, which is consistent with the goal of \emph{DQN} to maximize the final gain.

The above figures (Figs 2,3,5 and 8) can well reflect the convergence of the proposed scheme. The loss value keeps going down in Fig. 2, which indicates that our work is effective. Although affected by channel state, the energy consumption obtained by value function of different channel state in Fig. 5 all have the tendency to maintain the same value, which means that the diagram has a tendency to converge. As for Fig. 3 and Fig. 8, accuracy is another concrete form of value function, the tendency of the accuracy to remain stable in the subsequent part can well explain the convergence properties of the scheme.

\section{CONCLUSIONS}
In this paper, we have leveraged \emph{DQN} to explore the best trade-off between local and global updates with a given resource budget.  Thanks to the DT that can sensitively capture the dynamic changes of the network, the proposed scheme can adaptively adjust the aggregation frequency according to the channel state. Furthermore, an asynchronous federated learning architecture based on node clustering has been designed to eliminate the straggler effect,  which is more suitable for heterogeneous scenarios. The numerical results show that the proposed scheme is superior to the benchmark scheme in terms of learning accuracy, convergence rate and energy saving.

\section*{ACKNOWLEDGMENT}
This work was supported by the European Union's Horizon 2020 research and innovation programme under the Marie Skłodowska-Curie grant agreement No 824019, in part by the National Natural Science Foundation of China (61771373), in part by Special Funds for Central Universities Construction of
World-Class Universities (Disciplines), and in part by China 111 Project (B16037).

\end{document}